\documentclass[sigconf, screen]{acmart}
\usepackage{multirow}
\usepackage{algorithmicx}
\usepackage{algpseudocode}
\usepackage{algorithm}
\usepackage{enumitem}

\usepackage{color}
\AtBeginDocument{%
  }

\renewcommand\footnotetextcopyrightpermission[1]{}

\acmSubmissionID{2025-2297}

\begin{document}

\title[MARL-MambaContour]{MARL-MambaContour: Unleashing Multi-Agent Deep Reinforcement Learning for Active Contour Optimization in \\ Medical Image Segmentation}



\author{
    Ruicheng Zhang\textsuperscript{1*},
    Yu Sun\textsuperscript{1*},
    Zeyu Zhang\textsuperscript{2},
    Jinai Li\textsuperscript{1},
    Xiaofan Liu\textsuperscript{1},
    AU HOI FAN\textsuperscript{1},
    \\
    Haowei Guo\textsuperscript{1},
    Puxin Yan\textsuperscript{1†}
    \\[1.5ex] 
    \textsuperscript{1}Sun Yat-sen University \quad
    \textsuperscript{2}The Australian National University
    \\[1ex] 
    \textsuperscript{*}Equal contribution \quad
    \textsuperscript{†}Corresponding author
}

\renewcommand{\shortauthors}{R. Zhang, Y. Sun, et al.}









\begin{abstract}

We introduce MARL-MambaContour, the first contour-based medical image segmentation framework based on Multi-Agent Reinforcement Learning (MARL). Our approach reframes segmentation as a multi-agent cooperation task focused on generate topologically consistent object-level contours, addressing the limitations of traditional pixel-based methods which could lack topological constraints and holistic structural awareness of anatomical regions. Each contour point is modeled as an autonomous agent that iteratively adjusts its position to align precisely with the target boundary, enabling adaptation to blurred edges and intricate morphologies common in medical images. This iterative adjustment process is optimized by a contour-specific Soft Actor-Critic (SAC) algorithm, further enhanced with the Entropy Regularization Adjustment Mechanism (ERAM) which dynamically balance agent exploration with contour smoothness. Furthermore, the framework incorporates a Mamba-based policy network featuring a novel Bidirectional Cross-attention Hidden-state Fusion Mechanism (BCHFM). This mechanism mitigates potential memory confusion limitations associated with long-range modeling in state space models, thereby facilitating more accurate inter-agent information exchange and informed decision-making. Extensive experiments on five diverse medical imaging datasets demonstrate the state-of-the-art performance of MARL-MambaContour, highlighting its potential as an accurate and robust clinical application.

\end{abstract}

\vspace{-1em}




\begin{teaserfigure}
  \includegraphics[width=\textwidth]{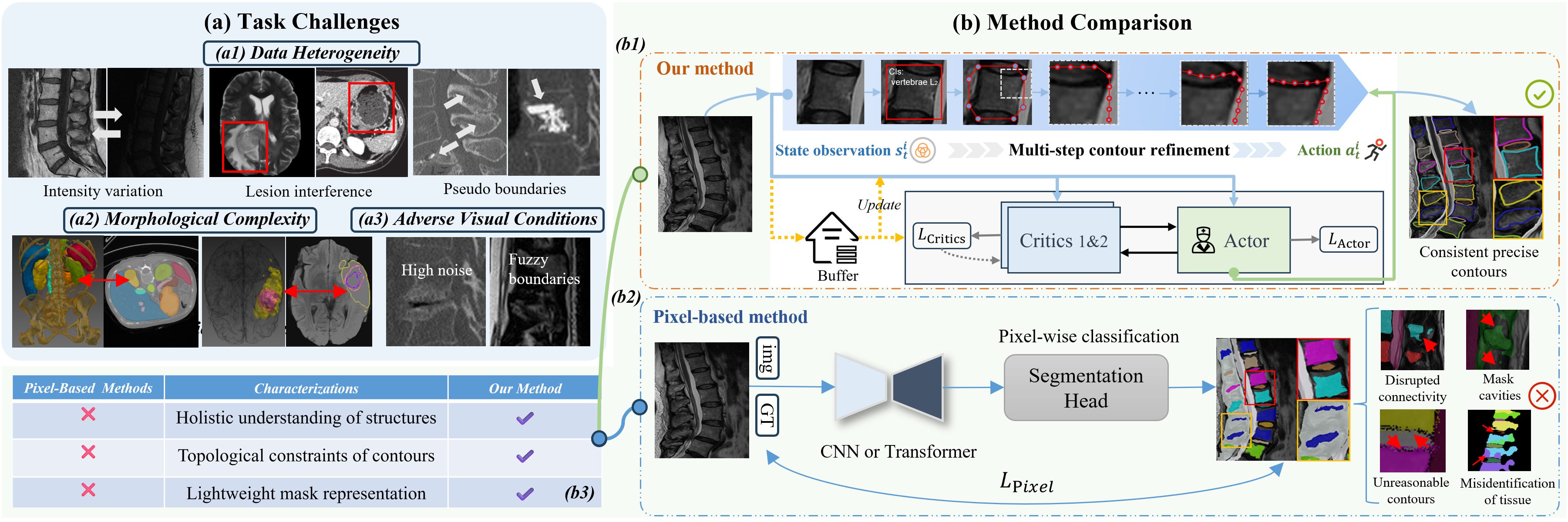}
  \vspace{-1em}
  \caption{(a) Task challenges in various aspects. (b) Design comparison between our method and pixel-based methods.}
  \label{fig:f1}
\end{teaserfigure}


\maketitle

\section{Introduction}

Medical image segmentation plays a critical role in diagnosis and treatment by delineating regions of interest (ROIs). In recent years, this field has progressed significantly, achieving increasingly accurate results through sophisticated neural network architectures and optimized training strategies \cite{RL_multi-step}. Most state-of-the-art segmentation methods, such as UNet \cite{unet} employ pixel-point-based classification techniques for segmentation across diverse modalities \cite{crisp-sam2}. However, despite their advancements, these methods often lack a holistic understanding of organ structures at the object level and fail to integrate topological constraints. As a result, they are vulnerable to challenges arising from medical images with heterogeneity, adverse visual conditions, and complex morphological structures (see Fig. \ref{fig:f1}(a)). This susceptibility frequently leads to erroneous outcomes, including mask cavities, pixel misclassification, disrupted connectivity, and irregular boundaries (see Fig. \ref{fig:f1}(b2)).

In contrast, the segmentation mask can also be represented as a sparse array of structured polygonal vertices that delineate the object’s contour \cite{PolyFormer,deep_snake}. This structured format imposes topological constraints on the masks, enabling the generation of smooth, precise contours even in complex medical images featuring blurred boundaries and intricate backgrounds. These contour-based methods excel at generating complete object-level contours, inherently avoiding the errors of pixel-based segmentation techniques. However, effectively driving these structured sequences to achieve precise contour delineation remains challenging, with previous efforts yielding limited success \cite{PolyFormer}. Existing methods \cite{PolarMask,PolyTransformer,deep_snake,PolyFormer,admira} inadequately address the morphological diversity of contour points and adaptively adjust contour smoothness accordingly, potentially leading to over-smoothing in places with high boundary complexity or knotting in regions of lower complexity. 


\enlargethispage{1\baselineskip}

The iterative segmentation process employed by physicians in delineating ROIs provides a more systematic and scientifically grounded framework. They initially define a bounding box around the foreground, followed by precise adjustments to each contour segment based on its local contour morphology. This approach leverages existing contour features as prior knowledge, iteratively refining the segmentation results to achieve greater accuracy, which aligns more closely with human comprehension of anatomical structures \cite{RL_multi-step}. Inspired by this, our work models contour delineation as a multi-agent Markov Decision Process, where each contour point displacement emulates the brushstroke actions of a clinician during segmentation. By doing so, we introduce a novel perspective to medical image segmentation, integrating the strengths of topological awareness and adaptive learning to enhance both accuracy and robustness in challenging imaging scenarios.

We introduce \textbf{MARL-MambaContour}, a piexl-based multi-step method for medical image segmentation that employs \textbf{M}ulti-\textbf{A}gent \textbf{R}einforcement \textbf{L}earning (\textbf{MARL}) for contour optimization. MARL-MambaContour performs object-level contour prediction through a human doctor-like progressive refinement workflow, consisting of initial contour generation followed by iterative contour evolution. In MARL-MambaContour, each contour point is modeled as an agent that iteratively navigates the \textit{2D} image space toward the target boundary, utilizing current contour features as prior knowledge. The movement of agents in continuous space is modeled as a Markov Decision Process (MDP) and addressed using a contour-specific Soft Actor-Critic (SAC) algorithm \cite{SAC}. To enhance morphological awareness during contour evolution, we integrate an adaptive \textbf{E}ntropy \textbf{R}egularization \textbf{A}djustment \textbf{M}echanism (\textbf{ERAM}) into SAC framework. ERAM dynamically adjusts the entropy regularization coefficient according to a contour consistency index, striking an adaptive balance between exploration and contour smoothness. When smoothness and consistence is lacking, the action policy reduces random exploration; conversely, it increases exploration intensity.

Contour, as a data structure with bidirectional causality, aligns naturally with the long-sequence modeling capabilities of the Mamba \cite{mamba}, which inspire us to bulid contour evolution policy network on Mamba architecture. To better adapt to point sequence data, MARL-MambaContour employs a bidirectional scanning strategy for contour information aggregation and interaction. However, the memory compression mechanism of conventional Mamba may confuse the features between different contour points, thereby limiting the precision of evolution. To address this limitation, we introduce the \textbf{B}idirectional \textbf{C}ross-attention \textbf{H}idden-state \textbf{F}usion \textbf{M}echanism (\textbf{BCHFM}), which enhances information alignment among contour points within the hidden state. By explicitly defining the attention contribution of each point in the hidden states of the bidirectional branches, the BCHFM provides more accurate and effective information for contour agent policies.

This study is the first to conceptualize medical image segmentation as a multi-agent contour reinforcement learning problem. Our contributions can be summarized as follows:
\begin{enumerate}
    \item We propose MARL-MambaContour, a pioneering framework that models medical image segmentation as a multi-agent Markov Decision Process. By iteratively leveraging prior knowledge from previous contour features as state information to guide subsequent segmentation decisions, our method emulates the coarse-to-fine scientific approach employed by clinicians.
    \item To enhance the optimization of contour point agents, we introduce an adaptive contour consistency-based Entropy Regularization Adjustment Mechanism (ERAM) within the Soft Actor-Critic (SAC) framework, effectively balancing the exploration capabilities of contour points with the smoothness of the resulting contour.
    \item We develop a state-space policy network featuring the Bidirectional Cross-attention Hidden-state Fusion Mechanism (BCHFM), enabling efficient feature aggregation and information transfer among contour points while mitigating memory blur from SSM compression.
    \item Evaluations on five challenging datasets demonstrate that MARL-MambaContour outperforms current state-of-the-art pixel-based and contour-based segmentation methods, highlighting its potential as a robust clinical tool.
\end{enumerate}

\section{Related Work}

\subsection{ Medical Image Segmentation} 
In recent years, segmentation models based on pixel-level classification techniques have achieved remarkable success, establishing themselves as the predominant approaches in medical image segmentation. These methods, primarily based on U-Net variants, exhibit exceptional performance in segmenting diverse tissues, including the spine \cite{2022spine3d, 2023vertebrae}, heart \cite{2021cycoseg, 2022bmanet}, brain tumors \cite{2021medical_transformer, 2024medsem}, blood vessels \cite{20223d_vessel, 2022dual}, and polyps \cite{2023mtanet, 2025UM-net}. Current research efforts mainly focus on developing more dedicated network architectures, optimizing training strategies, and effectively integrating attention mechanisms. Despite their impressive performance, these models depend on independent pixel-wise predictions, which inherently lack topological constraints and a holistic, object-level understanding of the segmented structures. Consequently, they may be more susceptible to prevalent medical imaging challenges, including substantial noise, artifacts, and morphological variations, often resulting in unreasonable errors such as mask cavities, disrupted connectivity, and erroneous regional mergers.


\subsection{Contour-based Segmentation}
In these methods, object shapes are represented as sequences of vertices along their boundaries. These approaches can be broadly categorized into two types. The first, exemplified by snake algorithms \cite{deep_snake,gamed-snake,admira,PolyTransformer,BoundaryTransformer}, deform an initial contour to align with the target boundary. This approach benefits from object-level holistic perception and imposes strong topological constraints, thereby exhibiting robust performance. It excels at preserving smooth and plausible boundaries, particularly in challenging cases with blurred edges and high noise. The second category, represented by \cite{PolarMask,PolyFormer,Explicit_Shape}, focuses on directly regressing the coordinates of contour points from the original images. This approach achieves high precision while offering faster inference speeds, making it computationally efficient. Despite these strengths, research on contour-based segmentation remains underrepresented, particularly in the complex domain of medical image segmentation. In this field, challenges such as blurred boundaries, extensive morphological variation, and unpredictable lesions impede precise contour prediction, often resulting in overly smooth contours that inadequately fit the target.

\subsection{Deep Reinforcement Learning}
The integration of reinforcement learning with deep learning has yielded significant advancements across diverse applications. Deep Reinforcement Learning (DRL) \cite{DRL} demonstrates considerable potential in tackling complex visual tasks, including image classification \cite{RL-class}, object localization \cite{RL-Localization}, action recognition \cite{RL-action}, and face recognition \cite{RL-Face}. Some studies \cite{Interactive,MRL-Seg,RLSegNet} have also explored the application of DRL to medical image segmentation; however, these approaches are constrained by the inherent limitations of pixel-wise segmentation paradigms. Compared to pixel-based methods, the integration of DRL with iterative contour optimization offers a more natural alignment. Unlike traditional global optimization strategies, this sequential decision-making framework is more suitable for optimizing contours in shape perception and interactive iterations, effectively emulating the coarse-to-fine segmentation process employed by clinicians.

\begin{figure*}[!ht]
\centering
\setlength{\abovecaptionskip}{-0.02em}   
\includegraphics[width=1.0 \textwidth]{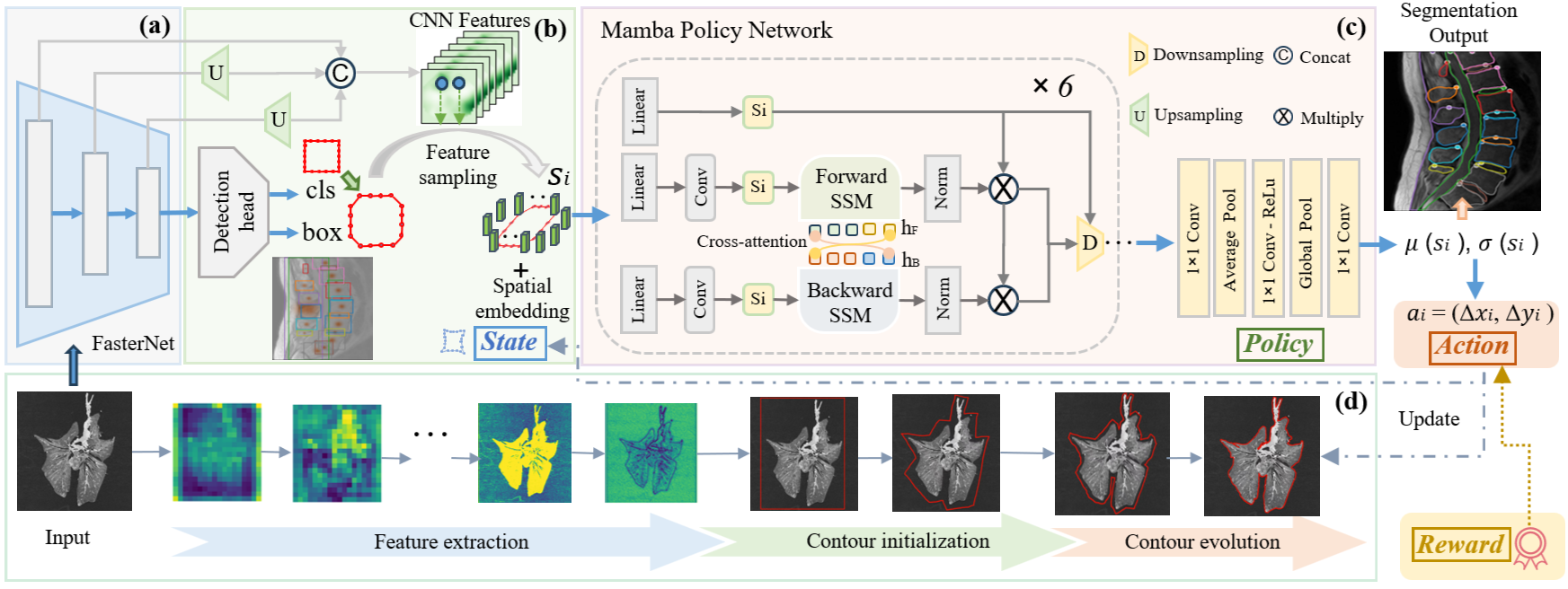}
\caption{ (a) Feature extraction backbone built on the FasterNet \cite{FasterNet}. (b) State space initialization. (c) Mamba policy network (Actor). (d) Segmentation pipeline for MARL-MambaContour.  
}
\label{fig:pipeline}  %
\vspace{-0.8em}
\end{figure*}

\section{Methodology}
\subsection{Overview}

MARL-MambaContour presents a novel approach to medical image segmentation by conceptualizing contour evolution as a multi-agent reinforcement learning (MARL) problem. As illustrated in Figure\ref{fig:pipeline}, the algorithm implements a systematic pipeline that mirrors the coarse-to-fine segmentation process employed by clinicians, comprising \textbf{\textit{initialization detection}} and \textbf{\textit{contour evolution}}. For a given medical image, a detector first generates a bounding box for each target organ, which is subsequently cropped into an octagon to serve as the initial contour. From each initial contour,  \( N \) points are uniformly sampled along its boundary. These points are treated as individual agents within a MARL framework, iteratively adjusting their positions to delineate the true organ boundary, as detailed in Section\ref{sec:marl}. Agent optimization is facilitated by a contour-tailored Soft Actor-Critic (SAC) algorithm \cite{SAC}, which balances exploration and conservatism via an adaptive Entropy Regularization Adjustment Mechanism (ERAM), as outlined in Section\ref{sec:sac}. In each iteration, agents perform position adjustments within a continuous action space, guided by a Mamba policy network enhanced with a Bidirectional Cross-Attention Hidden-State Fusion Mechanism (BCHFM), as described in Section\ref{sec:policy_net}. This iterative refinement persists until the contour converges to the target boundary, yielding precise and robust segmentation results suitable for clinical applications.

\subsection{Multi-Agent Contour Reinforcement Learning}
\label{sec:marl}
We propose a new contour optimization framework by modeling contour points as individual agents within a Multi-Agent Reinforcement Learning (MARL) paradigm. This design capitalizes on the MARL's capacity for decentralized decision-making and cooperative policy learning to tackle the complex task of boundary delineation in medical images. Unlike previous overall contour optimization methods \cite{deep_snake,gamed-snake}, where global contour adjustments may overlook local intricacies, our MARL-based method empowers each contour point with greater autonomous exploration capabilities. It iteratively refines its position while coordinating with others to achieve a globally coherent segmentation boundary.


The contour evolution process is formalized as a Markov Decision Process (MDP) within a multi-agent setting, where agents collaboratively learn optimal policies to balance local precision and global contour consistency. Below, we elaborate on the state space, action space, and reward function.

\noindent \textbf{State Space}

The state space for the \( i \)-th agent, which corresponds to the \( i \)-th contour point, is defined as a composite vector that integrates both spatial and contextual information:
\begin{equation}
s_i = \left[ (x_i, y_i), f_i, \text{Embedding}\{ (x_j, y_j) \}_{j \in \mathcal{N}_i} \right],
\end{equation}
where \( (x_i, y_i) \) represents the agent's current coordinates in the image plane, \( f_i \) denotes the local features extracted from a region centered at \( (x_i, y_i) \), and \( \text{Embedding}\{ (x_j, y_j) \}_{j \in \mathcal{N}_i} \) encapsulates the sinusoidal position encoding of neighboring agents within a predefined adjacency set \( \mathcal{N}_i \), which explicitly incorporates the spatial arrangement of multiple agents into the state.

The local features \( f_i \) are extracted using an Inception module \cite{Inception} from feature maps shared with the detector:
\begin{equation}
f_i = \mathcal{F}_{\text{Inception}}(I_f; x_i, y_i, r),
\end{equation}
where \( I_f \) represents the feature maps output by the feature extraction network, and \( r \) specifies the radius of the local region. These features are subsequently processed by the Mamba policy network, which adopts a shared parameter architecture to model interactions among agents. This shared learning mechanism promotes information exchange across the contour, enabling agents to effectively integrate local observations with global dynamics and thereby enhancing the overall coherence of the contour evolution process.

\noindent \textbf{Action Space} 

The action space is continuous, defined as the displacement of each contour point in the \( x \) and \( y \) directions:  
\begin{equation}
a_i = (\Delta x_i, \Delta y_i) \in \mathbb{R}^2.
\end{equation}
To maintain stability and coordination, we impose a bounded constraint $\| a_i \| \leq 25$ to control the maximum step size. This regularization prevents excessive deviations that could disrupt inter-agent alignment.

\noindent \textbf{Reward Function} 

The reward function is engineered to incentivize both individual accuracy and collective performance with dual focus on local and global objectives. It comprises distinct four components:

1. \textbf{\textit{Initialization Reward}}: In MARL-MambaContour, an object detector generates bounding boxes for each organ as initial contours. At this stage, we reward the alignment between the initial bounding box and the ground truth using the Dice coefficient: $\text{Dice}_{\text{init}} = \frac{2 |B_{\text{init}} \cap B_{\text{gt}}|}{|B_{\text{init}}| + |B_{\text{gt}}|}, $ where \( B_{\text{init}} \) is the region enclosed by the initial bounding box, and \( B_{\text{gt}} \) is the ground truth mask. The reward is defined as:
\begin{equation}
r_{\text{init}, i} = w_0 \cdot \text{Dice}_{\text{init}},
\end{equation}
where \( w_0 \) is a weighting factor. This term incentivizes the detector to produce high-quality initial contours, facilitating subsequent refinement.

2. \textbf{\textit{Region Overlap Reward}}: This component drives agents to collectively align the contour-enclosed region with the target mask, leveraging the mean Intersection over Union (mIoU): $\text{mIoU} = \frac{1}{N} \sum_{k=1}^{N} \frac{|A_k \cap B_k|}{|A_k \cup B_k|}$, where \( A_k \) is the region enclosed by the current contour for the \( k \)-th object, \( B_k \) is the ground truth mask, and \( N \) is the number of objects. The reward is defined as the incremental improvement:  
\begin{equation}
r_{\text{region}, i} = w_1 \cdot \left( \text{mIoU}^{(t)} - \text{mIoU}^{(t-1)} \right),
\end{equation}
where \( w_1 \) is a weighting factor, and \( t \) denotes the time step. By tying individual rewards to a global metric, this term encourages cooperative behavior, as each agent’s action impacts the shared mIoU.

3. \textbf{\textit{Boundary Alignment Reward}}: This reward emphasizes local precision, employing the mean Boundary F-score (mBoundF) \cite{boundf} to evaluate edge fidelity:
\begin{equation}
\text{mBoundF} = \frac{1}{5} \sum_{n=1}^{5} \text{mDice}_n (\partial A_n, \partial B_n),
\end{equation}
where \( \text{mDice} = \frac{1}{N} \sum_{i=1}^{N} \frac{2 |A_i \cap B_i|}{|A_i| + |B_i|} \), \( \partial A_n \) and \( \partial B_n \) are the boundary pixels of the predicted and ground truth regions, and \( \text{mDice}_n \) is the Dice coefficient across five distance thresholds. The reward is:
\begin{equation}
r_{\text{boundary}, i} = w_2 \cdot \left( \text{mBoundF}^{(t)} - \text{mBoundF}^{(t-1)} \right),
\end{equation}
with \( w_2 \) as a weighting factor. This term drives agents to refine their positions near the target boundary, enhancing local detail capture.

4. \textbf{\textit{Cooperative Regularization Term}}: To enforce multi-agent coordination, we introduce a regularization term based on contour smoothness:
\begin{equation}
r_{\text{coop}, i} = - w_3 \cdot \sum_{j \in \mathcal{N}_i} \| (x_i + \Delta x_i, y_i + \Delta y_i) - (x_j + \Delta x_j, y_j + \Delta y_j) \|_2^2,
\end{equation}
where \( w_3 \) is a weighting factor, and the summation penalizes large positional disparities between neighboring agents post-action. This term ensures contour coherence, mitigating irregularities often overlooked by single-agent methods.

The total reward for each agent combines these components:\
\begin{equation}
r_i = r_{\text{init}, i} + r_{\text{region}, i} + r_{\text{boundary}, i} + r_{\text{coop}, i}.
\end{equation}

These reward functions play distinct roles during different phases of contour evolution. During the early contour evolution, the region overlap rewards facilitate rapid convergence through agent cooperation. As the contour approaches the target boundary, the boundary alignment and cooperative terms take precedence to refine the segmentation result, ensuring a balance between global alignment and local boundary precision.

\vspace{-0.5em}
\subsection{Contour-specific Soft Actor-Critic}
\label{sec:sac}
The Soft Actor-Critic (SAC) algorithm introduces the concept of maximum entropy on the basis of maximizing cumulative future rewards, effectively enhancing the robustness and exploratory capabilities of agents in continuous action spaces. In this framework, contour points are rewarded for the randomness of their actions (i.e., higher entropy), preventing premature convergence to local optima during evolution. This mechanism is particularly beneficial in mitigating the interference of blurred boundaries and artifacts in medical images. Nevertheless, achieving a balance between the exploratory behavior of individual contour points and the overall robustness of the contour poses a significant challenge. Since the entropy regularization coefficient is often determined based on empirical human judgment, an excessively small value may result in overly smooth contours due to limited exploration, while an excessively large value can lead to jagged or self-intersecting contours.

To address this challenge, we propose a contour consistency-based \textbf{E}ntropy \textbf{R}egularization \textbf{A}djustment \textbf{M}echanism (ERAM), specifically tailored for contour optimization in medical image segmentation. ERAM dynamically adjusts the entropy regularization coefficient based on the contour’s consistency, ensuring an optimal trade-off between exploration and contour smoothness, thereby improving the overall quality of segmentation in complex medical imaging scenarios.

Specifically, we define a contour consistency index \( C \), which quantifies the variability in both distances and curvatures between consecutive contour points:
\begin{equation}
C = \lambda_1 \cdot \text{Var}\left( \{ d_{i, i+1} \mid i = 1, \dots, N-1 \} \right) + \lambda_2 \cdot \text{Var}\left( \{ \kappa_i \mid i = 1, \dots, N \} \right),
\end{equation}
where \( d_{i, i+1} \) is the Euclidean distance between adjacent points \( i \) and \( i+1 \), \( \kappa_i \) is the local curvature at point \( i \), computed based on neighboring points, \( N \) is the total number of contour points, and \( \lambda_1=0.1 \) and \( \lambda_2=0.5 \) are weighting parameters balancing the contributions of distance and curvature variability. The entropy regularization coefficient \(\alpha\) is then adaptively adjusted as:
\begin{equation}
\alpha = \alpha_0 \cdot \frac{1}{1 + \beta \cdot C},   
\end{equation}
where \(\alpha_0\) is the baseline entropy coefficient, and \(\beta\) is a learnable hyperparameter controlling the sensitivity to contour consistency (initialized at 0.05). When \( C \) is large (indicating an inconsistent or irregular contour), \(\alpha\) decreases, reducing exploration to prioritize contour smoothness. Conversely, when \( C \) is small (indicating a smooth contour), \(\alpha\) increases, encouraging exploration for finer boundary adjustments.

The SAC objective is formulated as:
\begin{equation}
J(\pi) = \mathbb{E} \left[ \sum_{t=0}^T \gamma^t \left( r_t + \alpha_t \mathcal{H}(\pi(\cdot|s_t)) \right) \right],
\end{equation}
where \(\gamma\) is the discount factor, \( r_t \) is the reward at time step \( t \), \(\mathcal{H}\) is the entropy of the policy, and \(\alpha_t\) is the time-varying entropy coefficient updated via ERAM.

To enhance the stability of value estimation, SAC employs a double Q-network structure, consisting of two independent Critic networks, \( Q_{\phi_1} \) and \( Q_{\phi_2} \). These networks are trained to estimate the state-action value function \( Q(s, a) \), and the target Q-value is computed using the minimum of the two estimates to mitigate overestimation bias:
\begin{equation}
Q_{\text{target}} = r + \gamma \left( \min_{j=1,2} Q_{\bar{\phi}_j}(s', a') - \alpha \log \pi(a'|s') \right),
\end{equation}
where \( a' \sim \pi(\cdot|s') \), and \( \bar{\phi}_j \) are the parameters of the target networks updated via soft synchronization.

The Actor network maps the state \( s_i \) of each contour point to its corresponding action \( a_i \), parameterized as a Gaussian distribution with mean \( \mu(s_i) \) and variance \( \sigma(s_i) \). We construct this network using the Mamba architecture, enhanced with the bidirectional hidden-state fusion mechanism (detailed in Section \ref{sec:policy_net}). The policy output is:
\begin{equation}
\mu(s_i), \sigma(s_i) = \pi_{\text{Mamba} \ \theta}(s_i),
\end{equation}
where \(\theta\) represents the trainable parameters of the Actor network.

The Critic networks, \( Q_{\phi_1} \) and \( Q_{\phi_2} \), estimate the state-action value function to evaluate the quality of actions taken by the Actor. We adopt a ResNet-18 backbone for processing the concatenated state-action pair:
\begin{equation}
Q_{\phi_j}(s_i, a_i) = \text{ResNet-18}(s_i, a_i), \quad j = 1, 2.
\end{equation}

\begin{small}
\begin{algorithm}[htbp]
\caption{MARL-MambaContour}
\label{alg:marl_mambasnake}
\begin{algorithmic}[1]
\State \textbf{Input:} Medical image \( I \), number of contour points \( N \), maximum iterations \( T \), batch size \( K \), hyperparameters \( \delta, w_1, w_2, w_3, \alpha_0, \beta, \gamma, \tau, \lambda_1, \lambda_2 \)
\State \textbf{Output:} Optimized contour points \( \mathcal{P}^{(T)} \)

\State Initialize Actor network \( \pi_\theta \) (Mamba-based), Critic networks \( Q_{\phi_1}, Q_{\phi_2} \), target networks \( Q_{\bar{\phi}_1}, Q_{\bar{\phi}_2} \), and replay buffer \( R \)
\State Use a detector to generate bounding boxes \( B \) for each organ in image \( I \)
\State Initialize contour points \( \mathcal{P}^{(0)} = \{p_1^{(0)}, p_2^{(0)}, \dots, p_N^{(0)}\} \) by uniformly sampling \( N \) points along the boundary of \( B \)
\State Extract initial states \( s_i^{(0)} = [(x_i^{(0)}, y_i^{(0)}), f_i^{(0)}, \Delta p_i^{(0)}] \) for each point

\For{step \( t = 1 \) to \( T \)}
    \State Compute contour consistency index \( C^{(t)} \) using current points \( \mathcal{P}^{(t)} \)

    \State Update entropy coefficient \( \alpha^{(t)} \) using ERAM:
    \[
    \alpha^{(t)} = \alpha_0 \cdot \frac{1}{1 + \beta \cdot C^{(t)}}
    \]

    \For{each agent \( i = 1 \) to \( N \)}
        \State Sample action \( a_i^{(t)} \sim \pi_\theta(\cdot | s_i^{(t)}) \) using Mamba policy network
        \State Update point position: \( p_i^{(t+1)} = p_i^{(t)} + a_i^{(t)} \), with \( \| a_i^{(t)} \|_2 \leq \delta \)
        \State Compute new state \( s_i^{(t+1)} \) for updated position \( p_i^{(t+1)} \)
        \State Compute reward \( r_i^{(t)} \):
        \[
        r_i^{(t)} = r^{(t)}_{\text{init}, i} + r^{(t)}_{\text{region}, i} + r^{(t)}_{\text{boundary}, i} + r^{(t)}_{\text{coop}, i}
        \]
        \State Store transition \( (s_i^{(t)}, a_i^{(t)}, r_i^{(t)}, s_i^{(t+1)}) \) in replay buffer \( R \)
    \EndFor

    \For{mini-batch \( k = 1 \) to \( K \)}
        \State Sample \( N \) transitions \( \{ (s_i, a_i, r_i, s_i') \}_{i=1,\dots,N} \) from \( R \)
        \State Compute target value \( y_i \) for each transition:
        \[
        y_i = r_i + \gamma \left( \min_{j=1,2} Q_{\bar{\phi}_j}(s_i', a_i') - \alpha^{(t)} \log \pi_\theta(a_i' | s_i') \right), \quad a_i' \sim \pi_\theta(\cdot | s_i')
        \]
        \State Update Critic networks \( Q_{\phi_1}, Q_{\phi_2} \) by minimizing the loss:
        \[
        L_j = \frac{1}{N} \sum_{i=1}^N \left( y_i - Q_{\phi_j}(s_i, a_i) \right)^2, \quad j = 1, 2
        \]
        \State Update Actor network \( \pi_\theta \) by minimizing the policy loss:
        \[
        L_\pi(\theta) = \frac{1}{N} \sum_{i=1}^N \left( \alpha^{(t)} \log \pi_\theta(a_i | s_i) - \min_{j=1,2} Q_{\phi_j}(s_i, a_i) \right)
        \]
    \EndFor

    \State Update target networks using soft updates:
    \[
    \bar{\phi}_1 \leftarrow \tau \phi_1 + (1 - \tau) \bar{\phi}_1, \quad \bar{\phi}_2 \leftarrow \tau \phi_2 + (1 - \tau) \bar{\phi}_2
    \]
\EndFor

\State \textbf{Return:} Optimized contour points \( \mathcal{P}^{(T)} \)

\end{algorithmic}
\end{algorithm}
\end{small}

\subsection{Mamba Policy Network}
\label{sec:policy_net}

The Mamba policy network is designed to map the state \( s_i \) of each contour point to its corresponding action \( a_i \), modeled as a Gaussian distribution with mean \( \mu(s_i) \) and variance \( \sigma(s_i) \). This network employs a bidirectional scanning strategy with forward and backward branches to capture bidirectional dependencies in contour data \cite{vm_unet}. The network architecture comprises six sequential processing layers, each featuring Mamba blocks based on the Selective State Space Model (SS2D) \cite{mamba_comprehensive}, downsampling, and feature concatenation. This configuration adeptly captures long-range dependencies and multi-scale features vital for precise contour refinement. 

The SS2D operation within each Mamba block follows the recursive formulation:
\begin{equation}
h_t = A h_{t-1} + B x_t, \quad y_t = C h_t + D x_t,
\end{equation}
where \( A, B, C, D \) are learnable parameters, \( h_t \) denotes the hidden state at time step \( t \), \( x_t \) is the input, and \( y_t \) is the output. This operation compresses global agent information into the hidden running state \( h_t \) ~\cite{mamba_comprehensive}, with \( \|A\| < 1 \)  regulating the decay of early input influence. This compression mechanism obscures feature memory \cite{mim_huyin} among distinct contour points, hindering precise decision-making by independent agents.

To address this limitation, we propose a Bidirectional Cross-attention \cite{huyin_naochuxue} Hidden-state Fusion Mechanism (BCHFM) to facilitate precise information alignment among contour points in the hidden state, effectively mitigating the obscures memory issue in Mamba. As a contour point's state is primarily influenced by its local neighbors rather than distant points, BCHFM adopts a windowed approach instead of employing a computationally expensive global cross-attention Specifically, the hidden state sequences \( h_{\text{fwd}} \) and \( h_{\text{fwd}} \) are partitioned into non-overlapping local windows of size \(w\). Cross-attention is then performed independently within each corresponding pair of windows. This strategy reduces the computational complexity from \(O(N^2)\) to a much more efficient \(O(N \cdot w)\) while preserving the most relevant contextual information.The fusion within each partition follows:

\vspace{-1.3em}
\begin{align}
\text{Attn}_{\text{fwd} \to \text{bwd}}^{\text{local}} &= \text{softmax}\left( \frac{(h_{\text{fwd}}^{\text{local}} W_Q) (h_{\text{bwd}}^{\text{local}, T} W_K)}{\sqrt{D}} \right) h_{\text{bwd}}^{\text{local}} W_V, \\
\text{Attn}_{\text{bwd} \to \text{fwd}}^{\text{local}} &= \text{softmax}\left( \frac{(h_{\text{bwd}}^{\text{local}} W_Q) (h_{\text{fwd}}^{\text{local}, T} W_K)}{\sqrt{D}} \right) h_{\text{fwd}}^{\text{local}} W_V, \\
h_{\text{fused}}^{\text{local}} &= \text{Attn}_{\text{fwd} \to \text{bwd}}^{\text{local}} + \text{Attn}_{\text{bwd} \to \text{fwd}}^{\text{local}},
\end{align}
where \( h^{\text{local}} \in \mathbb{R}^{w \times D} \) represents the hidden states within a window, and \( W_Q, W_K, W_V \) are shared projection matrices. The resulting fused states from all windows are then concatenated to form the final fused state \( h_{\text{fused}} \). Furthermore, since the number of contour points is capped at 128 in our implementation, the computational overhead of this partitioned fusion strategy is negligible.


The fused state \( h_{\text{fused}} \) is integrated into the SS2D updates of both branches using a learnable gating parameter \( \gamma \). 

\vspace{-1.3em}
\begin{align}
h_t^{\text{fwd}} &= A_{\text{fwd}} h_{t-1}^{\text{fwd}} + B_{\text{fwd}} x_t + \gamma \cdot h_{\text{fused}}, \\
h_t^{\text{bwd}} &= A_{\text{bwd}} h_{t+1}^{\text{bwd}} + B_{\text{bwd}} x_t + \gamma \cdot h_{\text{fused}},
\end{align}
where \( A_{\text{fwd}}, B_{\text{fwd}} \) and \( A_{\text{bwd}}, B_{\text{bwd}} \) are branch-specific parameters, and \( \gamma \) adjusts the influence of \( h_{\text{fused}} \).

The resulting feature representation \( h_{\text{final}} \) is processed by a Multi-Layer Perceptron (MLP) to derive the action distribution parameters:
\vspace{-1.3em}
\begin{equation}
\mu(s_i), \sigma(s_i) = \text{MLP}(h_{\text{final}}).
\end{equation}

The BCHFM provides precise, localized bidirectional context, enhancing decision-making for each contour point. It efficiently expands the receptive field to relevant neighbors, promotes effective collaboration among contour agents, and enables more accurate and reliable boundary detection without incurring significant computational costs.

\section{Experiments}
\begin{figure*}[!ht]
\centering
\setlength{\abovecaptionskip}{-0.02em}   
\includegraphics[width=1.0 \textwidth]{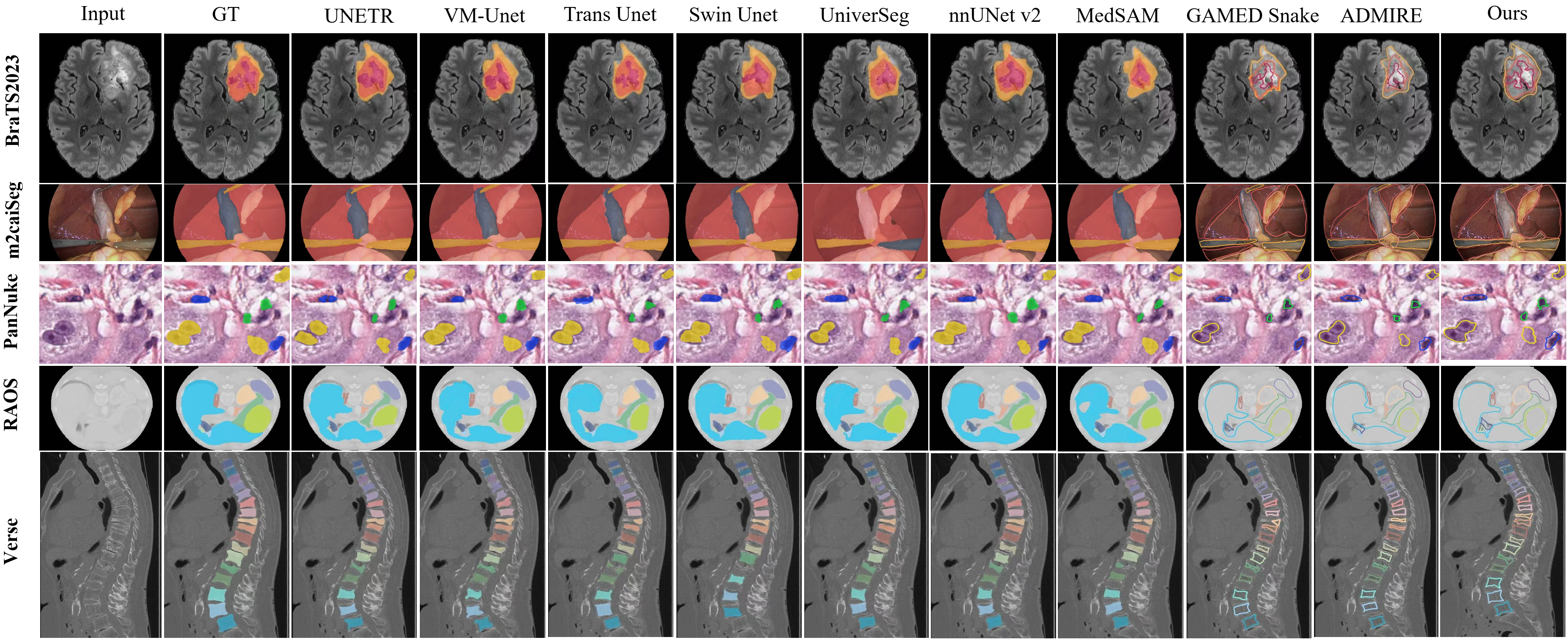}
\caption{ Qualitative comparison of segmentation results between MARL-MambaContour and other methods.
}
\label{fig:res}  %
\vspace{-0.8em}
\end{figure*}

\begin{table*}[h!]
\centering
\renewcommand{\arraystretch}{1}  
\setlength{\tabcolsep}{4pt}  
\resizebox{\textwidth}{!}{
\begin{tabular}{l|ccc|ccc|ccc|ccc|ccc}
\hline
\multirow{2}{*}{\textbf{Model}} & \multicolumn{3}{c|}{\textbf{BraTS2023 \cite{BraTS2023}}} & \multicolumn{3}{c|}{\textbf{Verse \cite{verse}}} & \multicolumn{3}{c|}{\textbf{RAOS \cite{RAOS}}} & \multicolumn{3}{c|}{\textbf{m2caiSeg \cite{m2caiseg}}} & \multicolumn{3}{c}{\textbf{PanNuke \cite{PanNuke}}}\\ \cline{2-16}
 & \textbf{mIoU} & \textbf{mDice} & \textbf{mBoundF}  & \textbf{mIoU} & \textbf{mDice} & \textbf{mBoundF} & \textbf{mIoU}  & \textbf{mDice} & \textbf{mBoundF}  & \textbf{mIoU} & \textbf{mDice} & \textbf{mBoundF}  & \textbf{mIoU} & \textbf{mDice} & \textbf{mBoundF} \\ \hline
nnUNet V2 \cite{nnunet} & 84.60 & 91.00 & 88.27 & 84.12 & 91.99 & 87.86 & 84.25 & 92.49 & 90.43 & 66.00 & 75.37 & 79.83 & 84.58 & 92.40 & 90.97 \\
UNETR \cite{unetr} & 84.17 & 91.71 & 87.45 & 83.28 & 90.14 & 86.08 & 83.95 & 92.61 & 90.62 & 66.18 & \textbf{76.69} & 80.30 & 84.09 & 92.29 & 91.03 \\
Trans Unet \cite{2021transunet} & 82.34 & 89.75 & 86.94 & 82.77 & 88.65 & 85.16 & 82.79 & 89.48 & 86.82 & 62.31 & 72.34 & 75.71 & 83.34 & 91.17 & 89.95 \\
Swin Unet \cite{swinunet} & 83.99 & 90.86 & 87.05 & 82.41 & 88.67 & 87.15 & 83.31 & 91.31 & 89.55 & 65.94 & 74.06 & 81.14 & 84.88 & 92.68 & 90.73 \\
VM-Unet \cite{vm_unet} & 83.12 & 90.03 & 86.47 & 81.64 & 87.24 & 86.58 & 82.89 & 91.06 & 89.47 & 64.30 & 73.19 & 78.01 & 83.56 & 92.68 & 89.68 \\
UniverSeg \cite{UniverSeg} & 82.07 & 88.77 & 84.14 & 80.08 & 87.19 & 84.43 & 81.60 & 88.93 & 88.38 & 59.06 & 69.88 & 56.57 & 82.47 & 88.70 & 89.27 \\
MedSAM \cite{2024medsem} & 84.46 & 90.80 & 88.01 & 81.05 & 87.80 & 85.13 & 82.42 & 89.21 & 85.98 & 62.31 & 58.13 & 68.53 & 52.53 & 90.53 & 90.25 \\ \hline
Gamed Snake \cite{gamed-snake} & 82.53 & 89.71 & 87.24 & 80.64 & 87.82 & 87.50 & 82.42 & 57.21 & 52.19 & 59.88 & 71.96 & 76.80 & 83.07 & 90.87 & 89.44  \\
ADMIRE \cite{admira} & 80.28 & 87.71 & 86.73 & 79.81 & 87.33 & 86.48 & 80.34 & 87.25 & 86.17 & 62.36 & 73.52 & 50.75 & 82.76 & 89.35 & 88.83 \\ \hline
\textbf{Ours} & \textbf{85.28} & \textbf{91.74} & \textbf{91.56} & \textbf{86.42} & \textbf{94.32} & \textbf{92.84} & \textbf{85.65} & \textbf{93.58} & \textbf{92.30} & \textbf{66.81} & {76.27} & \textbf{81.97} & \textbf{86.90} & \textbf{94.68} & \textbf{92.21} \\
\hline    
\end{tabular}
}
\caption{\textbf{Quantitative results.} All metrics are reported as \% values. Bold values indicate the best results in the table.}
\label{tab:comparison}
\vspace{-0.3in}
\end{table*}

\subsection{Experiment Configurations}
\textbf{Datasets}\quad To assess the efficacy of the MARL-MambaContour framework across diverse medical imaging modalities and anatomical structures, we conducted experiments on five challenging segmentation datasets: BraTS2023 (brain tumors, MRI) \cite{BraTS2023}, VerSe (spine, CT) \cite{verse}, RAOS (abdomen, CT) \cite{RAOS}, m2caiSeg (abdomen, endoscopic images.) \cite{m2caiseg}, and PanNuke (cells, microscopy) \cite{PanNuke}. These datasets present numerous anatomical categories, substantial morphological variations, severe pathological conditions, and blurred boundaries, making them well-suited to test the robustness and generalizability of our multi-agent contour evolution approach. Additional details are available in the supplementary materials.

\noindent \textbf{Evaluation Metrics}\quad We adopt three established metrics widely used in medical image segmentation to evaluate performance: mean Intersection over Union (mIoU), mean Dice Similarity Coefficient (mDice), and mean Boundary F-score (mBoundF). These are defined as: 
\vspace{-1.5em}
\begin{equation}
\begin{gathered}
\text{mIoU} = \frac{1}{N} \sum_{i=1}^{N} \frac{|A_i \cap B_i|}{|A_i \cup B_i|}, \ \ \
\text{mDice} = \frac{1}{N} \sum_{i=1}^{N} \frac{2 |A_i \cap B_i|}{|A_i| + |B_i|}, \\
\text{mBoundF} = \frac{1}{5} \sum_{n=1}^{5} \text{mDice}_n (\partial A_n, \partial B_n),
\end{gathered}
\end{equation}
where \( A_i \) represents the ground truth mask for the \( i \)-th organ, \( B_i \) denotes the predicted segmentation, and \( \partial A_n \) and \( \partial B_n \) are the boundary pixels of the ground truth and predicted masks, respectively, with \( n \) ranging from 1 to 5 pixels to account for varying boundary widths. These metrics collectively measure region overlap (mIoU and mDice) and boundary precision (mBoundF). These metrics collectively evaluate segmentation accuracy by measuring both region overlap (mIoU and mDice) and boundary precision (mBoundF), offering a robust evaluation of the model’s capability.

\noindent \textbf{Implementation Details}\quad The MARL-MambaContour framework employs the Mamba policy network as the Actor and two ResNet-18-based  \cite{resnet} Critic networks within the SAC algorithm. The feature extraction backbone, FasterNet \cite{FasterNet}, pre-trained on ImageNet \cite{ImageNet} and fine-tuned for each dataset, generates feature maps to guide contour evolution. The detection head, configured similarly to YOLO-v8 \cite{YOLOv8}, generates bounding boxes for each target organ, from which an octagonal initial contour is derived by selecting quartile points along each edge. Configuration parameters for the SAC algorithm are detailed in the supplementary materials. The Actor network (Mamba-based) and Critic networks (ResNet-18) are optimized using the AdamW optimizer with an initial learning rate of \(1\times10^{-4}\), which decays to \(1\times10^{-6}\) via cosine annealing over 200 epochs. Five-fold cross-validation is conducted on each dataset, with all experiments performed on two NVIDIA A100 GPUs.

\noindent \textbf{Comparing Experiments}\quad We compare MARL-MambaContoure with two categories of state-of-the-art methods: (1) \textbf{\textit{pixel-based segmentation models}}, including nnUNet v2 \cite{nnunet}, UNETR \cite{unetr}, TransUNet \cite{2021transunet}, SwinUNet \cite{swinunet}, VM-UNet \cite{vm_unet}, UniverSeg \cite{UniverSeg}, and MedSAM \cite{2024medsem}, and (2) \textbf{\textbf{contour-based segmentation models}}, such as ADMIRE \cite{admira} and GAMED-Snake \cite{gamed-snake}.

\vspace{-1em}
\subsection{Quantitative Results}

\subsubsection{\textbf{Overall}}
Table~\ref{tab:comparison} presents a comprehensive quantitative evaluation across five diverse datasets, confirming the consistent superiority of MARL-MambaContour over state-of-the-art pixel-based and contour-based methods.

Our approach achieves top performance in region overlap metrics (mIoU and mDice) across most datasets. This advantage mainly attributes to our MARL strategy that enables each contour point (agent) to iteratively refine its position using local image cues and collaborative signals. The Mamba policy network, enhanced by the Bidirectional Cross-Attention Hidden-State Fusion Mechanism (BCHFM), further improves this process by facilitating the aggregation of bidirectional contextual information, allowing agents to make informed decisions regarding their optimal placement relative to neighboring points and underlying image features.

Notably, the most significant advantage of MARL-MambaContour lies in its superior contour quality, as evidenced by the mBoundF metric. Our method effectively establishes topological constraints between contour points, promoting the generation of consistent, high-quality edges and inherently avoiding the jagged or discontinuous edges observed in pixel-wise methods. The substantial improvement in mBoundF across all datasets (e.g., +3.29\% on BraTS2023 and +4.98\% on Verse compared to the strongest competitor) is  attributed to the proposed Entropy Regularization Adjustment Mechanism (ERAM). By dynamically balancing exploration and robustness based on current contour consistency, ERAM enables MARL-MambaContour to precisely adapt to diverse boundary conditions, including both highly irregular tumor boundaries in BraTS2023 and smooth vertebral borders in Verse.
\vspace{-1em}

\begin{table}[htbp]
\centering
\scalebox{0.75} {
\begin{tabular}{lcccccc}
\toprule
\textbf{Models} & \textbf{Artery} & \textbf{Fat} & \textbf{Gallbladder} & \textbf{Intestine} & \textbf{Liver} & \textbf{Upperwall} \\
\midrule
nnUNet V2  &  29.41 &  57.90 &  69.23 & 88.28 &  63.12 & 86.07 \\
UNETR      &  31.57 &  59.74 &  71.22 & 88.13 &  62.44 & \textbf{86.36} \\
Trans Unet &  22.13 &  45.35 &  52.45 & 68.35 &  50.55 & 76.65 \\
Swin Unet  &  27.34 &  57.97 &  57.73 & 84.03 &  61.04 & 81.85 \\
VM-Unet  &  26.41 &  57.03 &  56.97 & 82.94 &  60.57 & 81.63 \\
MedSAM     &  24.22 &  51.64 &  52.47 & 88.27 &  61.06 & 77.68 \\
ADMIRE     &  26.38 &  55.51 &  54.94 & 92.87 &  62.28 & 80.82 \\
Gamed Snake&  28.40 &  57.46 &  68.12 & 87.64 &  62.75 & 85.03 \\
\midrule
\textbf{Ours} &  \textbf{37.68} &  \textbf{64.21} &  \textbf{72.48} & \textbf{89.35} &  \textbf{65.24} & 86.14  \\
\bottomrule
\end{tabular} 
}
\caption{Dice coefficients (\%) for segmentation performance of MARL-MambaContour. The best Dice score for each class for each model is highlighted in bold.}
\label{tab:m2caiSeg}
\vspace{-2em}
\end{table}

\subsubsection{\textbf{Considering substructures}}
Table~\ref{tab:m2caiSeg} presents a detailed performance evaluation of MARL-MambaContour across individual substructures within the challenging m2caiSeg endoscopic dataset.

Our approach consistently achieves the highest Dice scores for most substructures, particularly those with complex characteristics. For instance, its superior performance on 'Fat' tissue (64.21\% Dice, a +4.47\% improvement over the next best method) underscores the framework’s robustness in handling structures with diffuse or ambiguous boundaries, where traditional pixel-based methods often struggle. Additionally, our method demonstrates a notable advantage in segmenting small structures such as 'Artery' (37.68\% Dice, a +6.11\% improvement over the next best method), generating approximate contours upon successful identification despite minor limitations in contour precision.

This strong performance in challenging scenarios stems from the decentralized yet collaborative nature of the multi-agent reinforcement learning (MARL) approach. Each contour point agent refines its policy based on local state representations and inter-agent information exchange, allowing the contour to effectively navigate complex configurations, differentiate adjacent tissues, and accurately delineate individual substructures despite significant appearance variations or partial occlusions.

\begin{figure*}[!th]
\centering
\setlength{\abovecaptionskip}{-0.02em}   
\includegraphics[width=0.75\textwidth]{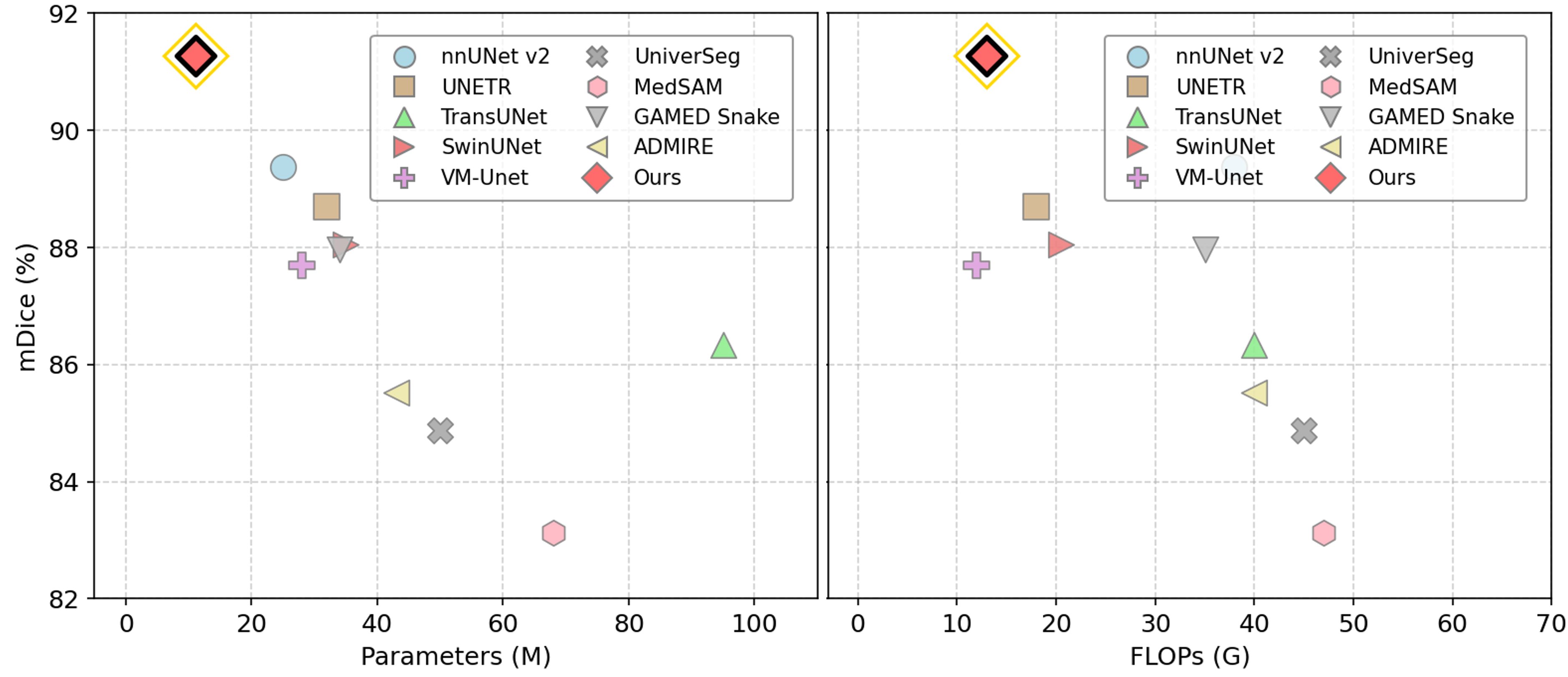}
\caption{  Model efficiency comparison.  }
\label{fig:eff}  %
\vspace{-0.5em}
\end{figure*}

\subsection{Qualitative Results}

Figure~\ref{fig:res} illustrates segmentation performance on representative slices from five datasets. Pixel-based approaches, such as MedSAM and TransUNet, struggle with challenging structures like tumor-affected regions (BraTS2023), densely packed vertebrae (VerSe), and overlapping nuclei clusters (PanNuke).
These challenges manifest as inaccurate boundary delineation, pixel misclassifications, and jagged contours. In contrast, MARL-MambaContour consistently generates coherent contours across diverse scenarios, effectively mitigating these errors. Furthermore, compared to other contour-based methods, MARL-MambaContour exhibits superior adaptability to complex morphologies. For instance, it accurately delineates concave liver regions (RAOS) and jagged tumor boundaries (BraTS2023), underscoring the effectiveness of our multi-agent contour evolution strategy.



\subsection{Ablation Study}
To comprehensively evaluate the efficacy of the reinforcement learning strategy, the contributions of key components and the influence of parameter configurations in the MARL-MambaContour framework, we conduct ablation studies on the RAOS dataset. 

\vspace{-0.5em}
\subsubsection{\textbf{Effectiveness of the Reinforcement Learning Strategy}}
To validate the effectiveness of our MARL-based contour optimization, we compare MARL-MambaContour against a baseline method similar to~\cite{deep_snake}, which directly supervises contour points using a distance loss to target boundary positions. As shown in Table~\ref{tab:ablation_rl}, the MARL approach significantly outperforms the supervised baseline, improving mIoU by 4.26\%, mDice by 4.02\%, and mBoundF by 4.16\%. These gains are attributable to our contour-specific SAC algorithm, which leverages multi-agent collaboration and iterative refinement to adaptively optimize contour points, enhancing both global coherence and local precision compared to static supervision.
\vspace{-1.8em}
\begin{table}[htbp]
\centering
\scalebox{0.75}{
\begin{tabular}{l|ccc}
\toprule
\textbf{Method} & \textbf{mIoU (\%)} & \textbf{mDice (\%)} & \textbf{mBoundF (\%)} \\
\midrule
Supervised Baseline & 82.15 & 89.96 & 88.61 \\
MARL-MambaContour & \textbf{85.65} (\textcolor{red}{$\Uparrow$4.26\%}) & \textbf{93.58} (\textcolor{red}{$\Uparrow$4.02\%}) & \textbf{92.30} (\textcolor{red}{$\Uparrow$4.16\%}) \\
\bottomrule
\end{tabular}
}
\caption{Effectiveness of Reinforcement Learning Strategy.}
\label{tab:ablation_rl}
\end{table}

\vspace{-2em}
\subsubsection{\textbf{Contributions of Key Components}}
We further investigate the individual and combined contributions of the Entropy Regularization Adjustment Mechanism (ERAM) and the Bidirectional Cross-attention Hidden-state Fusion Mechanism (BCHFM). Results are presented in Table~\ref{tab:ablation_components}, with the following observations:
\begin{itemize}
    \item \textit{ERAM Enhances Contour Accuracy}: Incorporating ERAM significantly improves mBoundF by 7.80\% (from 84.85\% to 91.47\%). This improvement stems from ERAM’s adaptive entropy regularization, which dynamically balances the exploration ability and robustness of the agent, ensuring high-quality contours even in regions with blurred boundaries or complex morphologies.
    \item \textit{BCHFM Boosts Overall Performance}: Adding BCHFM alone increases mIoU by 3.03\%, mDice by 4.20\%, and mBoundF by 4.84\%. This enhancement is due to BCHFM’s ability to aggregate bidirectional contextual information, improving coordination among contour points and capturing long-range dependencies critical for accurate boundary delineation.
    \item The full model, integrating both ERAM and BHFM, achieves the highest performance, demonstrating their complementary roles in optimizing contour evolution.
\end{itemize}

\vspace{-1em}
\begin{table}[htbp]
\centering
\scalebox{0.75}{
\begin{tabular}{l|ccc}
\toprule
\textbf{Configuration} & \textbf{mIoU (\%)} & \textbf{mDice (\%)} & \textbf{mBoundF (\%)} \\
\midrule
Baseline (w/o ERAM \& BCHFM) & 80.78 & 87.32 & 84.85 \\
+ ERAM & 83.79 (\textcolor{red}{$\Uparrow$3.73\%}) & 90.17 (\textcolor{red}{$\Uparrow$3.26\%}) & 91.47 (\textcolor{red}{$\Uparrow$7.80\%}) \\
+ BCHFM & 83.23 (\textcolor{red}{$\Uparrow$3.03\%}) & 90.99 (\textcolor{red}{$\Uparrow$4.20\%}) & 88.96 (\textcolor{red}{$\Uparrow$4.84\%}) \\
Full Model (ERAM + BCHFM) & \textbf{85.65}(\textcolor{red}{$\Uparrow$6.03\%}) & \textbf{93.58}(\textcolor{red}{$\Uparrow$7.19\%}) & \textbf{92.30}(\textcolor{red}{$\Uparrow$8.78\%}) \\
\bottomrule
\end{tabular}
}
\caption{Contributions of Key Components.}
\label{tab:ablation_components}
\vspace{-2em}
\end{table}

\vspace{-0.8em}
\subsubsection{\textbf{Parameter Configurations}}

We investigate the influence of varying the number of contour points \( N \) on segmentation performance (in Table~\ref{tab:point_num}). The configuration with \( N = 128 \) achieves the optimal balance between accuracy and training stability. Fewer points (\( N = 32 \)) result in insufficient contour resolution, while a higher number of points (\( N = 256 \)) leads to a slight performance decline, likely due to optimization challenges associated with an excessive number of agents.

Additionally, we analyze the impact of evolution iterations (in Table~\ref{tab:iterations}). Five iterations represent the optimal trade-off between performance and computational cost. Beyond five iterations, performance only has a slight improvement, yet computational overhead significantly increases.
\vspace{-1.0em}
\begin{table}[htbp]
\centering
\scalebox{0.80}{
\begin{tabular}{l|cccc}
\toprule
\textbf{Number of Contour Points} & 32 & 64 & 128 & 256 \\
\midrule
\textbf{mIoU (\%)} & 73.26 & 80.62 & \textbf{85.65} & 82.06 \\
\textbf{mDice (\%)} & 79.81 & 85.34 & \textbf{93.58} & 88.42 \\
\textbf{mBoundF (\%)} & 60.37 & 75.25 & \textbf{92.30} & 90.12 \\
\bottomrule
\end{tabular}
}
\caption{Performance of Different Contour Point Numbers.}
\label{tab:point_num}
\vspace{-1em}
\end{table}

\vspace{-0.7em}
\begin{table}[htbp]
\centering
\scalebox{0.80}{
\begin{tabular}{l|ccccc}
\toprule
\textbf{Evolution Iterations} & 3 & 4 & 5 & 6 & 7 \\
\midrule
\textbf{mIoU (\%)} & 81.54 & 83.17 & 85.65 & 85.67 & 85.67 \\
\textbf{mDice (\%)} & 87.57 & 91.07 & 93.58 & 93.63 & 93.61 \\
\textbf{mBoundF (\%)} & 86.73 & 89.23 & 92.30 & 92.38 & 92.37 \\
\bottomrule
\end{tabular}
}
\caption{Performance Across Different Evolution Iterations.}
\label{tab:iterations}
\vspace{-1em}
\end{table}

\subsubsection{\textbf{Robustness to Contour Initialization}}
As a two-stage method, the performance of our algorithm depends on the precision of the initial contours provided by the detection stage. We therefore conduct a sensitivity analysis to assess the robustness of the contour evolution process to initialization errors.
The results, summarized in Table~\ref{tab:sensitivity_detailed}, demonstrate the strong resilience of our method. For both positional and scale perturbations of $\pm$10\%, the reduction in the Dice coefficient is less than 0.5\%. Even with larger perturbations of $\pm$20\%, the performance drop remains modest, with a Dice reduction of approximately 2\%.

\vspace{-1em}
\begin{table}[htbp]
\centering
\scalebox{0.90}{
\begin{tabular}{lcc}
\toprule
\textbf{Perturbation Type} & \textbf{Magnitude $\pm$10\%} & \textbf{Magnitude $\pm$20\%} \\
\midrule
Positional Shift & 0.27\%$\Downarrow$ & 1.88\%$\Downarrow$ \\
Scale Jitter     & 0.43\%$\Downarrow$ & 2.04\%$\Downarrow$ \\
\bottomrule
\end{tabular}
}
\caption{Sensitivity to bounding box initialization errors, showing mDice reduction on the Verse validation set  \cite{verse}.}
\label{tab:sensitivity_detailed}
\vspace{-1em}
\end{table}

Although the contour evolution process is robust to variations in the initial bounding box, a complete failure in the detection phase would prevent the evolution from starting. To alleviate this dependency, we initialize the contour evolution with ground-truth bounding boxes during training. This strategy ensures stable learning by mitigating error propagation from the detection module, thereby enhancing the overall robustness of the model.

\section{Limitation}
We discusses several limitations of our approach under specific scenarios: 
\textit{(1) Handling Instances with Holes}: While our MARL-MambaContour excels at delineating fine-grained boundary contours, it faces challenges when processing objects with internal holes. \textit{(2) Fine-Grained or Disconnected Structures}: The contour-based model exhibits reduced performance when segmenting extremely small objects (e.g., objects spanning only a few pixels) or structures with topologically disconnected boundaries. In such scenarios, pixel-based segmentation methods may outperform our approach due to their ability to handle fragmented or minute structures. \textit{(3) Dependence on Detection}: The success of the contour evolution process is contingent upon the initial detection of objects. If the detector fails to identify an object, the multi-agent evolution cannot proceed, leading to inevitable segmentation failure. These limitations highlight areas for future optimization and improvement in contour-based segmentation.

\vspace{-0.5em}
\section{Conclusion}

In this work, we present MARL-MambaContour, a novel approach to medical image segmentation that leverages Multi-Agent Reinforcement Learning (MARL) for contour-based optimization. Our method models each contour point as an independent agent within a cooperative MARL framework. This approach iteratively refines segmentation contours to fit the target boundaries, emulating the coarse-to-fine delineation process used by clinicians. The integration of an adaptive Entropy Regularization Adjustment Mechanism (ERAM) into the Soft Actor-Critic (SAC) algorithm enables a dynamic equilibrium between exploration and contour smoothness, facilitating precise boundary alignment in challenging medical images. Additionally, the Mamba-based policy network augmented with the Bidirectional Cross-attention Hidden-state Fusion Mechanism (BCHFM) enhances the capture of bidirectional dependencies and mitigates feature memory obscuration, leading to more accurate decision-making by the agents. Our method not only advances contour-based medical image segmentation but also underscores the transformative potential of MARL in addressing complex visual tasks, positioning it as a promising robust and precise clinical tool.

\bibliographystyle{ACM-Reference-Format}
\bibliography{sample-base}
\appendix
\clearpage
\setcounter{page}{1}
\begin{center}
    \textbf{\Huge Supplementary Material}
\end{center}
\section{SAC Parameter Settings}
\label{sec:sac_params_supp_table}

The hyperparameters for the Soft Actor-Critic (SAC) algorithm, employed for optimizing the contour point agents in MARL-MambaContour, are carefully configured as detailed in Table \ref{tab:sac_parameters}. These settings are chosen to ensure stable training and effective policy learning for the contour evolution task.

\begin{table}[htbp]
\centering
\caption{Hyperparameter settings for the Soft Actor-Critic (SAC) algorithm and related components used in MARL-MambaContour.}
\label{tab:sac_parameters}
\resizebox{\columnwidth}{!}{
\begin{tabular}{@{}llcl@{}}
\toprule
\textbf{Category} & \textbf{Parameter} & \textbf{Symbol} & \textbf{Value} \\
\midrule 
Core SAC & Discount Factor & \(\gamma\) & 0.99 \\
 & Target Smoothing Coefficient & \(\tau\) & 0.005 \\
 & Replay Buffer Size & - & \(1 \times 10^6\) \\
 & Batch Size & \(K\) & 128 \\
ERAM (Adaptive Entropy) & Baseline Entropy Coefficient & \(\alpha_0\) & 0.2 \\
 & Sensitivity Parameter & \(\beta\) & 0.5 \\
 & Distance Variance Weight & \(\lambda_1\) & 0.1 \\
 & Curvature Variance Weight & \(\lambda_2\) & 0.5 \\
Reward Function Weights* & Initialization Reward Weight & \(w_0\) & 0.5 \\
 & Region Overlap Reward Weight & \(w_1\) & 1.0 \\
 & Boundary Alignment Reward Weight & \(w_2\) & 1.5 \\
 & Cooperative Regularization Weight & \(w_3\) & 0.1 \\
Action Space & Max Displacement Step & \(\delta\) & 25 pixels \\
Networks \& Optimizer & Actor Network Architecture & - & Mamba-based with BCHFM \\
 & Critic Network Architecture & - & 2x ResNet-18 \\
 & Optimizer & - & AdamW \\
 & Initial Learning Rate & - & \(1 \times 10^{-4}\) \\
 & Final Learning Rate & - & \(1 \times 10^{-6}\) \\
 & LR Schedule & - & Cosine Annealing \\
Training Parameters & Training Epochs & - & 200 \\
 & Contour Points per Instance & \(N\) & 128 \\
 & Evolution Iterations per Prediction & \(T_{\text{evolve}}\) & 5 \\
\bottomrule
\multicolumn{4}{l}{\footnotesize *Reward weights were determined through experimental tuning.} \\
\end{tabular}
}
\end{table}

\section{Method for 3D Generalization}
Although Mamba Snake is fundamentally a 2D paradigm, we extend it to 3D segmentation by effectively leveraging inter-slice coherence. Our 2.5D adaptation, which processes volumes slice-by-slice, proceeds as follows:

\begin{enumerate}
    \item \textbf{Decomposition and Propagation:} The 3D volume is first decomposed into a sequence of 2D axial slices. The segmentation process is initiated on the central slice, which typically contains the most representative anatomical information. The segmentation then propagates bidirectionally towards the extremities of the volume.

    \item \textbf{Inter-Slice Contour Initialization:} To maximize efficiency, we employ a dual-mode initialization strategy for subsequent slices:
    \begin{itemize}
        \item For \textit{continuing structures} already present in the preceding slice, the previously predicted contour serves as a high-quality initialization. This enables convergence to a precise segmentation in a single evolution step.
        \item For \textit{emerging structures} appearing for the first time, an initial octagonal contour is generated from a detection box and is subsequently refined over five evolution steps to accurately delineate its boundary.
    \end{itemize}

    \item \textbf{Volume Reconstruction:} Finally, the resulting 2D segmentation masks are stacked to reconstruct the complete 3D volume.
\end{enumerate}

This slice-wise propagation strategy is highly efficient because the majority of structures persist across adjacent slices, requiring minimal computational overhead for refinement. Consequently, our approach achieves competitive accuracy while maintaining a significant advantage in inference speed over fully 3D architectures.

\section{Dataset Introduction}
\label{sec:datasets_supp_corrected}
This study employs five influential segmentation datasets to assess the performance of our model: BraTS2023 (brain tumors, MRI) \cite{BraTS2023}, VerSe (spine, CT) \cite{verse}, RAOS (abdomen, CT) \cite{RAOS}, m2caiSeg (abdomen, endoscopic images) \cite{m2caiseg}, and PanNuke (cells, microscopy) \cite{PanNuke}. These datasets cover a range of tissues, imaging modalities, and exhibit multi-scale structural heterogeneity, providing a comprehensive test for our method. Detailed descriptions are provided below.

\textbf{BraTS2023} \quad The BraTS2023 dataset \cite{BraTS2023} serves as a standard benchmark for evaluating algorithms on brain tumor segmentation using multi-modal MRI (T1, T1ce, T2, FLAIR). It includes hundreds of patient scans, each annotated for three distinct tumor sub-regions: the enhancing tumor (ET), the tumor core (TC, excluding edema), and the whole tumor (WT, including edema). The significant challenges posed by this dataset include the high degree of variability in tumor appearance, shape, size, and location across patients; the subtle intensity differences and indistinct boundaries between tumor sub-regions and healthy tissue; and the inherent class imbalance, particularly for the smaller enhancing tumor region. We process each 3D volume by densely slicing it into a sequence of 2D images along the sagittal plane and then perform a 5-fold cross-validation to ensure a robust evaluation of our model's performance.

\textbf{VerSe}\quad The VerSe dataset \cite{verse}, specifically VerSe'19 and VerSe'20, provides a large-scale resource for vertebra label and segmentation from clinical CT scans. It contains 374 scans from 355 patients, offering substantial diversity in anatomy and image quality. Annotations cover individual vertebrae from the cervical (C1) to lumbar (L5) regions, including common anatomical variations like T13 and L6 (labeled 1-24, with L6 as 25 and T13 as 28). Key difficulties arise from the morphological similarity of adjacent vertebrae, requiring precise localization, significant variations in CT resolution and field-of-view, and boundary ambiguities often caused by degenerative diseases (e.g., osteophytes) or close inter-vertebral spacing. Our experiments utilize mid-sagittal plane reconstructions derived from these CT volumes, evaluated via 5-fold cross-validation.

\textbf{RAOS}\quad The RAOS dataset \cite{RAOS} presents a realistic clinical scenario for segmenting multiple abdominal organs from CT scans. Containing 413 scans from two different institutions, it reflects typical variations in acquisition protocols and patient populations. The primary segmentation challenges stem from the often low soft-tissue contrast between adjacent organs (e.g., distinguishing bowel loops from surrounding mesentery or fat), the indistinct and complex boundaries where organs touch or overlap, the considerable inter-patient variability in organ shape, size, and relative positioning, and the common presence of imaging artifacts (e.g., streak artifacts from dense material) that can disrupt anatomical contours. We process each 3D volume by densely slicing it into a sequence of 2D images along the sagittal plane and used 5-fold cross-validation.

\textbf{m2caiSeg}\quad The m2caiSeg dataset \cite{m2caiseg}, derived from the MICCAI 2016 Surgical Tool Detection challenge videos, shifts the focus to segmentation within endoscopic surgical scenes. It provides pixel-level annotations for several classes, including organs (liver, gallbladder, etc.), surgical tools, and other elements like bile or blood, across 307 annotated frames (split into 245 train, 62 test). This domain presents unique and severe challenges: highly variable and often specular illumination, frequent occlusions by instruments or tissue, visual obstruction from smoke or fluids, significant deformation and irregular appearance of tissues during manipulation, and a dynamic, limited field of view. Standard image normalization is applied and the official data split is used for evaluation.

\textbf{PanNuke}\quad The PanNuke dataset \cite{PanNuke} targets the complex problem of nuclei instance segmentation and classification within digital histopathology images, crucial for computational pathology. It comprises thousands of annotated image patches extracted from 19 distinct tissue types, reflecting a broad range of cancerous and healthy tissue appearances. The substantial challenges include the extremely high density of nuclei in many tissues, leading to frequent touching and overlapping instances that require accurate separation. Additionally, there is vast variability in nuclear morphology (size, shape, texture) and staining intensity, both within a single tissue type and especially across different tissues. Complex background textures and potential staining artifacts further complicate the task of precise nuclei delineation. Color normalization and intensity standardization are performed, and evaluation uses 5-fold cross-validation.




\end{document}